\documentclass[final,12pt]{elsarticle}

\usepackage{makecell}

\usepackage{amssymb}
\usepackage{amsmath}

\usepackage{framed} 
\usepackage{multicol} 
\usepackage{nomencl} 
\makenomenclature
\usepackage{tabularx}
\renewcommand*\nompreamble{\begin{multicols}{2}}
\renewcommand*\nompostamble{\end{multicols}}
\usepackage{amsmath}
\usepackage{amssymb}
\usepackage{arydshln} 
\usepackage{amsthm}
\usepackage{algorithm}
\usepackage{longtable}
\usepackage{algorithmic}
\usepackage{graphicx}
\usepackage{adjustbox}
\usepackage{subfig}
\usepackage{color}
\usepackage{url}
\definecolor{lightblue}{RGB}{0,176,240}
\definecolor{purple}{RGB}{153,102,255}
\usepackage{amsmath}
\usepackage[scr=boondox,  
            cal=esstix]   
           {mathalpha}
\usepackage{algorithm}
\usepackage{float}
\usepackage{multirow}
\usepackage[utf8]{inputenc}
\usepackage{lineno}
\usepackage[colorlinks,linkcolor=blue, anchorcolor=blue, citecolor=blue]{hyperref}

\journal{XXXXXXXXXX}

\begin{document}

\begin{frontmatter}
\title{High-Quality Cloud-Free Optical Image Synthesis Using Multi-Temporal SAR and Contaminated Optical Data}

\author[label1]{Chenxi Duan\corref{cor2}}
\ead{c.duan@utwente.nl}

\address[label1]{Faculty of Geo-Information Science and Earth Observation (ITC), University of Twente, P.O. Box 217, 7500 AE Enschede, The Netherlands}

\cortext[cor2]{Corresponding author}

\begin{abstract}

\par Addressing gaps caused by cloud cover and the long revisit cycle of satellites is vital for providing essential data to support remote sensing applications. This paper tackles the challenges of missing optical data synthesis, particularly in complex scenarios with cloud cover. We propose CRSynthNet, a novel image synthesis network that incorporates innovative designed modules such as the DownUp Block and Fusion Attention to enhance accuracy. Experimental results validate the effectiveness of CRSynthNet, demonstrating substantial improvements in restoring structural details, preserving spectral consist, and achieving superior visual effects that far exceed those produced by comparison methods. It achieves quantitative improvements across multiple metrics: a peak signal-to-noise ratio (PSNR) of 26.978, a structural similarity index measure (SSIM) of 0.648, and a root mean square error (RMSE) of 0.050. 
\par Furthermore, this study creates the TCSEN12 dataset, a valuable resource specifically designed to address cloud cover challenges in missing optical data synthesis study. The dataset uniquely includes cloud-covered images and leverages earlier image to predict later image, offering a realistic representation of real-world scenarios. This study offer practical method and valuable resources for optical satellite image synthesis task.
\end{abstract}
\end{frontmatter}

\section{Introduction}

\par Remote sensing images have been extensively utilized across various fields such as disaster management \citep{nguyen2022improvement}, climate change \citep{yang2021evolution}, and urban planning \citep{wang2022unetformer}, demonstrating remarkable performance. The continuous development of algorithms has significantly enhanced their efficiency and effectiveness, further expanding their applications. Meanwhile, more satellites are being developed and launched to acquire more remote sensing data \citep{SONG2024104932}. Despite these advancements in both algorithms and satellite technology, challenges in obtaining reliable and high-quality data remain a fundamental limitation in many applications \citep{gonzalezcalabuig2025generative}. Missing or delayed data disrupts real-time monitoring and hinders temporal analysis in remote sensing. For instance, the absence of satellite images can lead to inaccurate disaster predictions and poorly planned evacuations \citep{zhu2022urban}. Additionally, delays in identifying affected areas slow down recovery efforts, such as resource distribution. This lack of timely information not only affects disaster response and recovery but also makes it challenging to fully evaluate vegetation growth and agricultural conditions \citep{christovam2021pix2pix}.

\par The causes of missing data can generally be attributed to several factors. Firstly, sensor malfunctions issues may occur unexpectedly, leaving gaps in the collected data. For instance, the Landsat 7 Scan Line Corrector (SLC) experienced a malfunction and caused stripping in subsequent images \citep{li2024improved}. Secondly, incomplete data collection can arise from satellite orbit limitations and data transmission disruptions between satellites and ground stations. Lastly, atmospheric conditions, such as cloud cover, can lead to incomplete images, particularly affecting optical images since synthetic aperture radar (SAR) images are not affected by weather conditions \citep{zeng2020towards}. These obstacles significantly hinder the accessibility and reliability of satellite-based data collection, further limiting various applications. Compared to SAR images, which can capture data in all-weather and all-day \citep{zeng2020towards}, optical images are highly affected by clouds, haze, and low sunlight, limiting reliability in cloudy or rainy regions \citep{sun2017haze}. Given the challenges posed by data loss in optical images, this paper specifically addresses the issue of missing optical data.

\par The issue of missing optical images can be addressed through two primary approaches: cloud removal and full-image generation. Cloud removal focuses on small-scale missing regions, which are often caused by localized clouds or sensor errors \citep{xu2022attention}. Full-image generation addresses scenarios where the entire image becomes unusable due to data transmission errors or extensive thick cloud coverage \citep{yang2022sar}. In such cases, synthesizing a complete optical image often rely on auxiliary data, such as multi-temporal optical images or SAR data \citep{dong2024integrating}. While both approaches aim to mitigate the issue of missing data, they differ significantly in their scope and application. Cloud removal can be seen as a specialized task within image generation, focusing on partial restoration rather than synthesizing an entire scene.

\par Based on the data modalities used, these methods can be classified into two main categories: single-source-dependent methods and multisource-dependent methods. Methods that rely on single-source non-optical data, commonly referred to as image translation, with SAR-to-optical translation being the most widely used technique \citep{yang2022sar}. Methods that depend on single-source optical image are referred to in this paper as optical self-restoration. Multisource-dependent methods integrate data from various sources, including multiple optical data, multiple non-optical data, and a combination of optical and non-optical data \citep{dong2024integrating}. These methods collectively referred to as image fusion. 

\par In the early stages, spatial statistical methods were employed for optical self-restoration task \citep{addink1999comparison, van2012remote}. These methods, however, often lead to reduced textural details and introduce smoothing effects. With the advancement of algorithms, cloud removal tasks have been addressed through an increasingly diverse range of techniques, including information cloning \citep{lin2012cloud} and deep learnig \citep{zhang2018missing}. These methods primarily leverage complementary information from images captured by the same satellite, either by utilizing cloud-free information within itself or integrating observations from different acquisition time. Multitemporal cloudy image series offer valuable complementary information. To better utilize the temporal context, certain methods employ longer temporal sequences of data to produce cloud-free image series \citep{duan2020thick,chen2024thick,zhang2021combined}. These methods are relatively efficient and capable of leveraging multitemporal information, making them well-suited for scenarios with consistent temporal data. However, these methods are highly sensitive to significant variations between images. When temporal changes are too pronounced, the data characteristics may fail to align with the prior knowledge used to constrain the target function, thereby limiting the method's effectiveness in accurately reconstructing cloud-free images. Due to the low temporal resolution of single-source data and the limited reference information it provides,  many methods integrate multiple optical data sources to reduce revisit intervals and generate cloud-free images under broader usability conditions \citep{wang2025mst, zhou2020reconstruction}.

\par SAR-optical translation involves converting SAR data into optical image, enabling the use of optical-like features in applications where optical data is unavailable or incomplete. \citep{merkle2017possibility} were the first to demonstrate the application of the Pix2Pix-based method for translating SAR images into grayscale optical image within a designated area. These methods primarily rely on various types of GAN architectures, such as CycleGAN \citep{fuentes2019sar}, S-CycleGAN \citep{wang2019sar}, and enhanced cGAN \citep{yang2022sar, doi2020gan, guo2021edge}, but the translation between different imaging mechanisms and the limited availability of reference information result in insufficient accuracy.

\par SAR-optical image fusion for image reconstruction combines SAR data with optical image to leverage their complementary characteristics, effectively reconstructing cloud-free images even in challenging weather or obstructed conditions. Residual neural networks were the first network applied in this field \citep{meraner2020cloud}, laying the foundation for further integration of global information \citep{xu2022glf}, which achieved improved results. However, these methods remain computationally intensive, limiting their applicability in real-world scenarios. Therefore, more efficient and rapid approaches have proposed \citep{mao2022cloud, duan2024feature, duan2024efficient}. Additionally, some scholars have introduced multitemporal image fusion methods that take both optical and SAR image sequences as inputs to produce cloud-free optical image serise \citep{xu2023multimodal}. Nevertheless, the complexity of these networks is significantly higher than single-date cloud removal. Meanwhile, certain researchers have suggested using a single temporal supplementary  image for time series fusion, which has demonstrated promising results while requiring fewer computational resources compared to long temporal sequence data \citep{dong2024integrating}. Even though it is more accurate than SAR-translation methods, it holds potential for improvement. For instance, the assumption that all optical images used for supplemental information are entirely cloud-free does not align with real-world conditions due to the inability to completely avoid clouds in optical image.

\par To address this issue, we have proposed a novel image synthesis method, called CRSynthNet (Cloud Resilient Synthesis Network). The contributions of this paper are listed as follows:
\par 1. We designed CRSynthNet to perform effectively even with cloud-contaminated temporal images, enhancing its practicality for real-world applications.
\par 2. We designed a FusionAttention mechanism to prioritize the fusion of data from two temporal phases, enhancing the network's ability to generate missing images effectively.
\par 3. We designed a novel Decoder that integrates features across multiple levels, by emphasizing critical information across spatial and channel domains, leading to improved accuracy and performance in image synthesis tasks.
\par 4. To evaluate the practical applicability of this method, we proposed a TCSEN12 (Two Temporal Sentinel-1 and Sentinel-2) dataset covering the area affected by the Zhengzhou 20-July-2021 flood. This dataset has large land cover change because of the flood, and it is suitable to evaluating the effectiveness of the methods.

\section{Dataset and Method}\label{sec: method}
\subsection{TCSEN12 Dataset}
\par To address the challenge of robustly synthesize of cloud-free optical images, we constructed a novel dataset focused on the Zhengzhou flood region. This dataset is designed to facilitate research in optical image synthesis and multi-temporal SAR-optical data fusion.
\par The Sentinel-1 and Senintel-2 images in this dataset is collected in Zhengzhou city and surrounding areas affected by the flood. Specifically, the dataset convered area is geographically bounded by longitude ranging from 112.0°E to 116.0°E and latitude spanning 33.0°N to 36.5°N. The images span the period from July 1, 2021, to July 31, 2021. Data preprocessing steps involve geospatial alignment, composite generation, and normalization. The generation of the 6-day composite aims to achieve more consistent temporal alignment, addressing the challenge of data originating from  different time points. Additionally, it helps reduce atmospheric interference, thereby enhancing the chances of obtaining lower cloud coverage images for training. 

\par Each data instance consists of one cloudy optical image from \textit{Time} 1, one SAR image from \textit{Time} 1 and \textit{Time} 2, and the reference near cloud-free optical image from \textit{Time} 2. The near cloud-free image is defined as having less than 5\% cloud coverage, as achieving completely cloud-free images is rarely feasible in real-world conditions.

\par Compared to the existing datasets, this dataset uniquely integrates cloudy optical images with multi-temporal SAR data, enabling enhanced modeling of real-world conditions such as cloud occlusion. The dataset spans the flood period, capturing notable changes in surface conditions, which makes it particularly suitable for evaluating the robustness of detection methods.

\subsection{Method}
\begin{figure}[h]
    \begin{minipage}{0.45\textwidth}
        \centering
        \includegraphics[width=\textwidth]{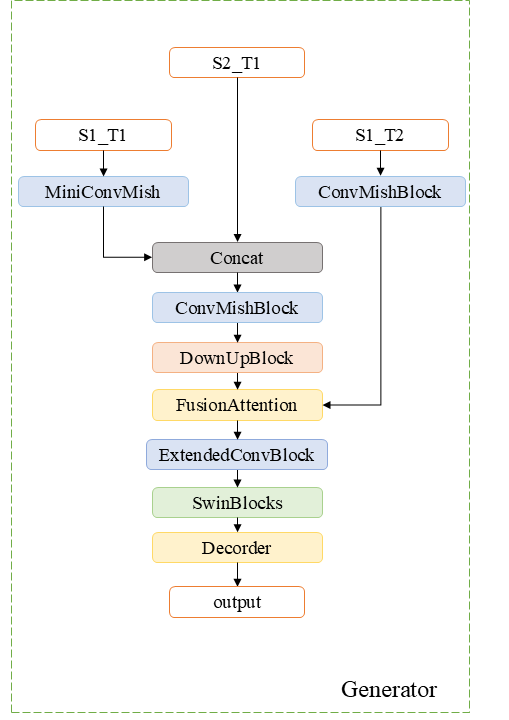}
        \caption{The architecture of the proposed generator. The detailed structure of each block is in Figure \ref{method_downup} to Figure \ref{method_decorder}.}
        \label{method_gen}
    \end{minipage}
    \hfill
    \begin{minipage}{0.45\textwidth}
        \centering
        \includegraphics[width=\textwidth]{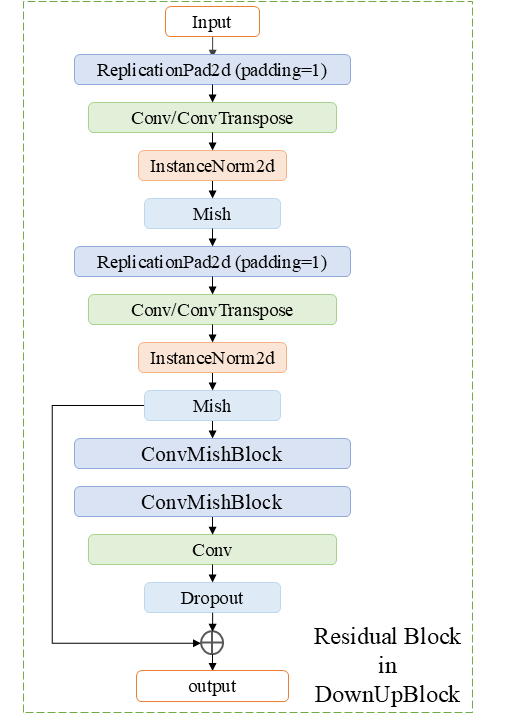}
        \caption{Detailed components of the DownUp block. The ConvMishBlock is illustrated in Figure \ref{method_Mish}.}
        \label{method_downup}
    \end{minipage}   
\end{figure}

\begin{figure*}[htbp]
\centering
\includegraphics[width=14cm]{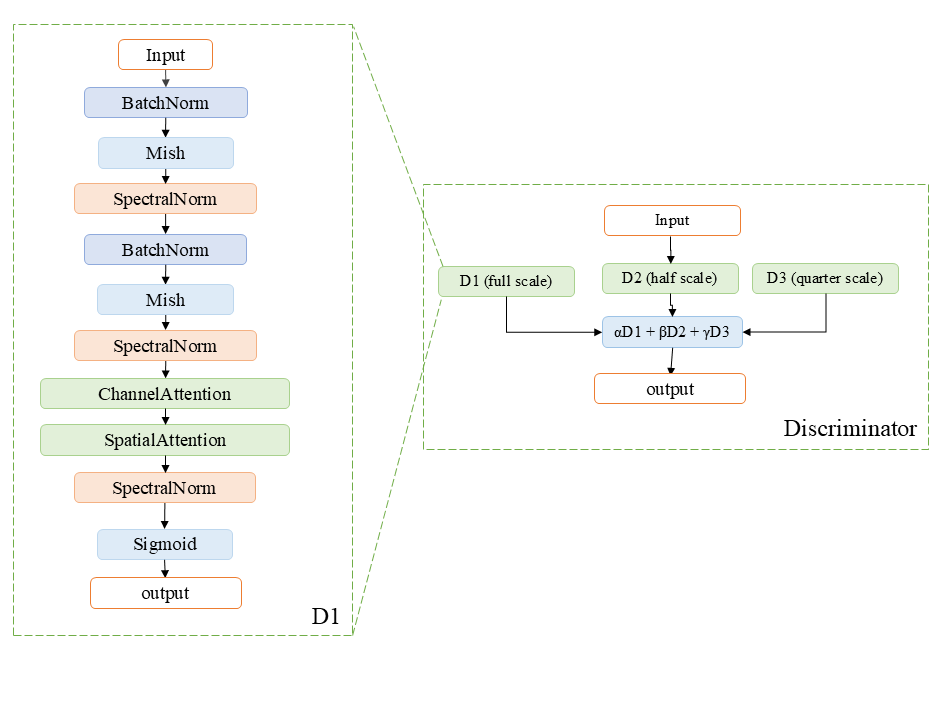}
\caption{The architecture of Discriminator with the structure of D1.}
\label{method_dis}
\end{figure*}

Our proposed GAN-based network, consisting of a generator and a discriminator, is designed to generate clear optical images that remain unaffected by cloud contamination in the input data. The generator produces synthetic images, while the discriminator differentiates between the generated images and actual real-world images (reference image), essentially helping the generator improve its output by providing feedback. The architecture of our proposed network, including both the generator and the discriminator, is detailed in Figure \ref{method_gen} to Figure \ref{method_dis}. In the following, we describe the generator, discriminator, and loss function in-depth.

\subsubsection{Generator}
The generator in our proposed network produces optical images for the desired time, regardless of whether the input images contain cloud contamination. It incorporates a sequence of Convolutional blocks, a DownUp block, a fusion attention block, a SwinBlock, and a Decoder. The generator's input consists of the Sentinel-1 images from both \textit{Time} 1 (S1\_T1) and \textit{Time} 2 (S1\_T2), along with the Sentinel-2 image from \textit{Time} 1 (S2\_T1). The processing of these three input images within the network is illustrated in Figure \ref{method_gen}. The key components of the Generator are as follows.

\begin{figure*}[htbp]
\centering
\includegraphics[width=14cm]{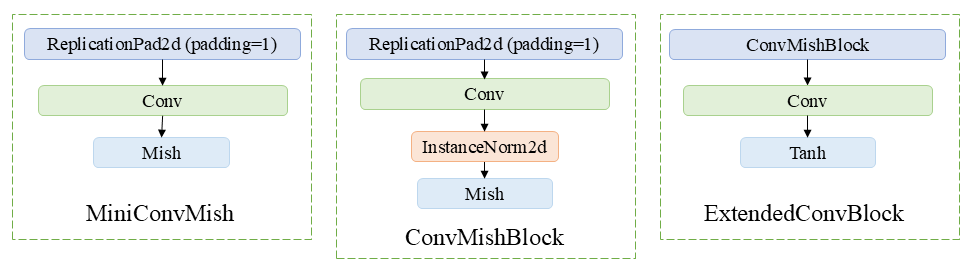}
\caption{The detailed design of three convolutional blocks.}
\label{method_Mish}
\end{figure*}

\begin{figure*}[htbp]
\centering
\includegraphics[width=14cm]{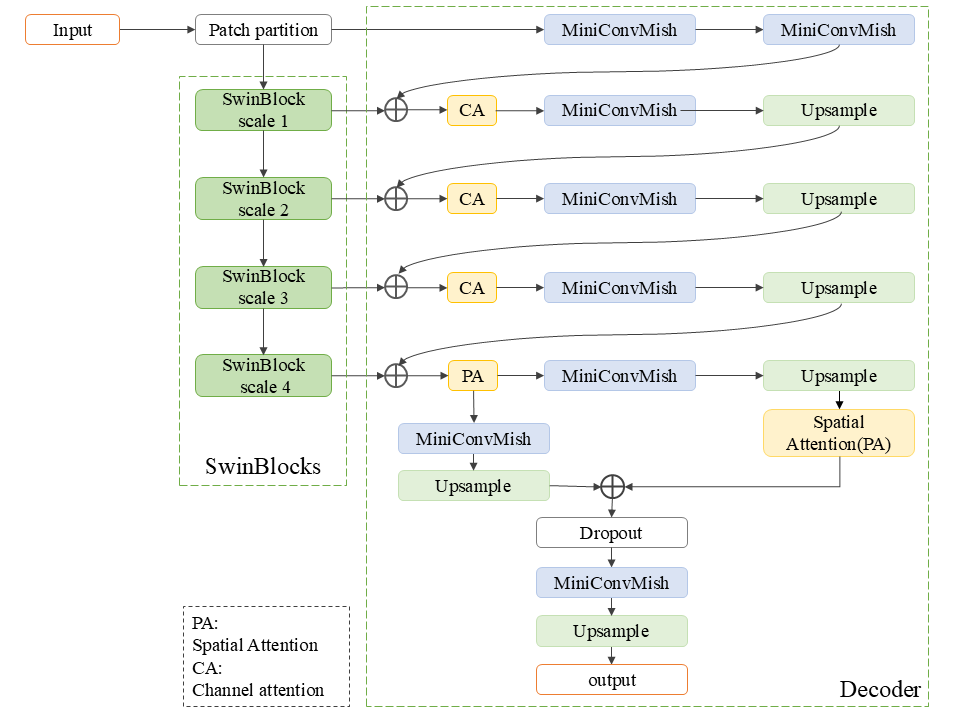}
\caption{The detailed architecture of Decoder.}
\label{method_decorder}
\end{figure*}

\begin{enumerate}
    \item \textbf{Convolutional Blocks}: The convolutional modules in our network are composed of three types (shown in Figure \ref{method_Mish}): MiniConvMish, ConvMishBlock, and ExtendedConvBlock. MiniConvMish consists of a replication padding layer with a padding size of 1, followed by a 3$\times$3 convolutional layer, and a Mish activation function. Based on the MiniConvMish module, ConvMishBlock adds an Instance Normalization layer to further normalise the feature maps. Building on the MiniConvMish, ExtendedConvBlock further includes a 1$\times$1 convolutional layer and a Tanh activation function for additional feature transformation.    
    \begin{figure*}[htbp]
    \centering
    \includegraphics[width=14cm]{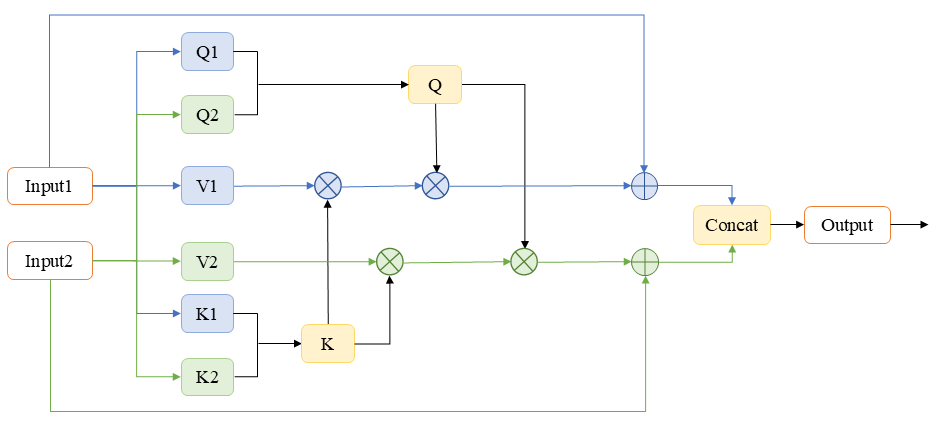}
    \caption{The specific architecture of FusionAttention modules. The blue arrows represent operations related to input1, while the green arrows indicate operations associated with input2.}
    \label{method_att}
    \end{figure*}
    \item \textbf{DownUpBlocks}: The DownUpBlock in our network comprises two down-sampling steps followed by two up-sampling steps, each performed by a designed residual block (Figure \ref{method_downup}). The input features of this block are initially down-sampled to reduce their dimensions and capture essential information. The features are subsequently up-sampled to restore their original dimensions, facilitating high-quality feature extraction. Each down-sampling and up-sampling operation employs convolutional layers, instance normalization, and Mish activation to ensure effective processing and transformation of the input data. Specifically, convolutional layers are applied during down-sampling, while deconvolutional layers (also known as transposed convolutional layers) are used during up-sampling. This block focuses primarily on the fusion of two data modalities, i.e., SAR and optical images, from time point T1, extracting the combined features.
    \item \textbf{FusionAttention}: This block is designed to incorporate self-attention mechanisms for enhancing feature fusion by applying joint attention mechanisms to the two input feature maps (Figure \ref{method_att}). It uses convolutional layers to generate query, key, and value representations for each input feature. The queries and keys from both inputs are then concatenated along the channel dimension, followed by $\mathcal{L}_2$ normalisation to ensure stability. Assuming that the normalised query and key tensors are represented as \(Q\) and \(K\), the attention is represented as:
    \begin{equation}
    \text{Attention}_1 = Q \cdot (K \cdot V_1), \quad \text{Attention}_2 = Q \cdot (K \cdot V_2),
    \end{equation}
    where \( V_1 \) and \( V_2 \) are the value tensors derived from the input feature maps.
    The final attention-enhanced outputs are weighted by the normalization term and integrated with the original inputs via a residual connection parameterized by \( \gamma \). This process is defined as:
    \begin{equation}
    y_1 = input_1 + \gamma \cdot \text{Attention}_1, \quad y_2 = input_2 + \gamma \cdot \text{Attention}_2,
    \end{equation}
    where:
    \begin{itemize}
    \item \( input_1 \) and \( input_2 \): Original input feature maps.
    \item \( \gamma \): A learnable scalar parameter controlling the influence of the attention-weighted components.
    \item \( \text{Attention}_1 \) and \( \text{Attention}_2 \): Attention-modulated feature maps derived from the value tensors \( V_1 \) and \( V_2 \).
    \end{itemize}
    The residual connection ensures that the attention-enhanced features contribute adaptively to the final outputs.
    \item \textbf{SwinBlocks}: In our network design, we adopt the scaled-up Swin Transformer architecture, utilizing its increased capacity and enhanced resolution to improve performance and efficiency \citep{liu2022swin2}. Specifically, we utilise four scaled-up Swinblocks to generate multi-scale features. This design provides several advantages, including hierarchical feature representation, and robust computation through scaled cosine attention. This design improves the network's ability to capture fine-grained details and handle complex patterns in high-resolution features.

    \item \textbf{Decoder}: The stucture of Decoder is illustrated in Figure \ref{method_decorder}. This block processes encoded feature maps from the SwinBlocks to fuse multi-scale features. It incorporates dropout to mitigate overfitting and utilizes convolutional and upsampling layers to enhance spatial resolution and produce the final output. Additionally, it leverages both channel and spatial attention mechanisms to improve feature representations. Channel attention is applied to features with low spatial resolution and multiple channels, whereas spatial attention is employed for features with higher spatial resolution and fewer channels.     
\end{enumerate}

\subsubsection{Discriminator}
\par The discriminator in a Generative Adversarial Network (GAN) functions as a binary classifier, distinguishing between real and generated data. Its primary purpose is to assess the generator's outputs and provide feedback on how closely they resemble the real data distribution. This adversarial training process encourages the generator to produce increasingly realistic outputs over time.
\par As shown in Figure \ref{method_dis}, the discriminator in this architecture is composed of three hierarchical subdiscriminators, i.e., D1, D2 and D3, each designed to handle features at different spatial resolutions. Specifically, the first subdiscriminator processes the input at its original resolution, while the second and third subdiscriminators process downsampled versions of the input using interpolation with scale factors of 0.5 and 0.25, respectively. These subdiscriminators share a unified structure, differing in the scales and the number of channels in the features they process. The architecture integrates batch normalization, Mish activation functions, and spectral normalization to ensure stable training and effective feature extraction. Furthermore, it employs channel and spatial attention mechanisms to refine feature representations. By selectively weighting important spatial or channel-wise regions, attention helps the discriminator better capture subtle differences and fine-grained details. This improves feedback to the generator, resulting in more realistic outputs. The outputs of these subdiscriminators are combined to form the final decision. This multi-scale design enables the discriminator to effectively capture features across varying resolutions, improving its capability to differentiate between real and generated images.

\subsubsection{Loss function}
The proposed network involves two competing neural networks: a generator ($G$), and a discriminator ($D$). Their losses, $loss\_G$ and $loss\_D$, play a crucial role in the training process by facilitating this competition. The generator loss ($loss\_G$) measures how well the generator creates realistic samples that can "fool" the discriminator. The discriminator loss ($loss\_D$) assesses the discriminator's ability to differentiate between real data and fake data. 

\par $loss\_G$ in this paper combines the least squares loss (\( L_{ls} \)) \citep{mao2017least} and Similarity Loss (\( L_{sim} \)) \citep{dong2024integrating}. This combined approach aims to optimise the generator's ability to produce outputs that not only closely resemble real data but also maintain high fidelity. The balance achieved by the weighting factor \( \lambda \) is critical in fine-tuning the quality and diversity of the generated outputs. The overall generator loss is shown as Equation~\ref{eq:loss_G}.

\begin{equation}
\begin{split}
loss\_G =  L_{sim}(prediction, reference) + \\
               \lambda L_{ls}(
                  D((prediction, \text{S}1\_\text{T}1, \text{S}1\_\text{T}2, \text{S}2\_\text{T}1), 1))              
\end{split}
\label{eq:loss_G}
\end{equation}
where S1\_T1, S1\_T2, and S2\_T1 respectively refer to Sentinel-1 images acquired at \textit{Time} 1 and \textit{Time} 2 as well as optical image obtained at \textit{Time} 1.

\par In Equation~\ref{eq:loss_G}, \( \text{L}_{\text{sim}} \) integrates Visual Geometry Group (VGG) Loss (\( L_{\text{VGG}} \)), Cosine Similarity Loss (\( L_{\text{CS}} \)), and MS-SSIM Loss (\( L_{\text{MS-SSIM}} \)) to comprehensively evaluate the similarity between the generated image and the reference image. This evaluation encompasses three distinct aspects: perceptual feature similarity, directional similarity of feature vectors, and structural similarity. The formulation of \( L_{sim} \) is presented as follows.  

\begin{equation}
\begin{split}
L_{sim} &= \alpha \cdot L_{\text{VGG}}(prediction, reference) \\
&\quad + \beta \cdot \big( 1 - L_{\text{CS}}(prediction, reference) \big) \\
&\quad + \gamma \cdot \big( 1 - L_{\text{MS-SSIM}}(prediction, reference) \big)
\end{split}
\label{eq:L_sim}
\end{equation}

\begin{itemize}
    \item \textbf{Perceptual Feature Similarity:} Achieved using \( L_{\text{VGG}} \) \citep{simonyan2015a}, this component measures the perceptual similarity between the generated and reference images by comparing high-level features extracted from a pre-trained VGG network.

\begin{equation}
L_\text{VGG} = \frac{1}{W_{i,j} H_{i,j}} \sum_{x=1}^{W_{i,j}} \sum_{y=1}^{H_{i,j}} \left( \phi_{i,j}(I_\text{ref})_{x,y} - \phi_{i,j}(I_\text{gen})_{x,y} \right)^2
\label{eq:vgg_loss}
\end{equation}

Where \( W_{i,j} \) and \( H_{i,j} \) represent the width and height of the feature map at layer \( \phi_{i,j} \) in the VGG network. \( \phi_{i,j} \) denotes the feature extractor at the \( i \)-th layer and \( j \)-th channel. \( I_\text{ref} \) is the reference image, and \( I_\text{gen} \) represents the generated image. 

\begin{equation}
L_{\text{CS}} = I - \frac{I_\text{gen} \cdot I_\text{ref}}{\|I_\text{gen}\| \|I_\text{ref}\|}
\label{eq:spectrum_loss}
\end{equation}

where $I$ denote a tensor assigned a constant value of 1.

\item \textbf{Structural Similarity:} MS-SSIM \citep{MS-SSIM} focuses on the structural resemblance between the generated and reference images across multiple scales, accounting for factors such as luminance, contrast, and local structure. It extends the original SSIM \citep{SSIM} by progressively downsampling the images and computing SSIM at various scales. This multi-scale approach allows MS-SSIM to capture differences across a range of spatial frequencies, providing a more robust and comprehensive evaluation of image quality.
    \begin{equation}
    {L}_\text{MS-SSIM} = 1 - \text{MS-SSIM}(I_\text{gen}, I_\text{ref}),
    \end{equation}
    where \( I_\text{gen} \) and \( I_\text{ref} \) represent the generated and reference images, respectively.
\end{itemize}

In Equation~\ref{eq:loss_G}, the formula for \( {L}_{\text{ls}} \) is given as:
\begin{equation}
{L}_{\text{ls}} = \frac{(D(G(S1\_T1, S1\_T2, S2\_T1)) - 1)^2}{2}
\label{eq:ls_loss}
\end{equation}

\par $loss\_D$, the discriminator loss, which evaluates the realness of generated samples and real data, plays a vital role in the adversarial training process. The Wasserstein GAN (WGAN) loss function \citep{pmlr-v70-arjovsky17a} with gradient penalty stabilises the training process and encourages convergence. Below, we detail its formulation and function.
\begin{equation}
loss\_D = L_\text{real} + L_\text{fake} + \lambda_\text{gp} \cdot L_\text{gp}
\label{eq:discriminator_loss}
\end{equation}

where \( {L}_\text{real} = -\mathbb{E} \left[ D(x_{\text{real}}) \right] \) represents the loss for real samples, while \( {L}_\text{fake} = \mathbb{E} \left[ D(x_{\text{fake}}) \right] \) is the loss for fake samples. Here, \( \mathbb{E} \) represents the expected value, \( {L}_\text{gp} \) is the gradient penalty term, and \( \lambda_\text{gp} \) is a weighting factor that controls the contribution of the gradient penalty. 
\par Gradient penalty is a crucial component for ensuring that the discriminator satisfies the Lipschitz constraint required by the Wasserstein GAN formulation. It involves generating interpolated samples between real and fake data, computing the gradients of the discriminator's output with respect to interpolated samples, and measuring the \( L_{2} \) norm of the gradients. The penalty term is defined as the squared difference between \( L_{2} \) norm and 1, which ensures the gradients are close to 1. By integrating this penalty into the loss function, the training process achieves improved stability and mitigates overfitting in the discriminator.

\section{Experiments}
\subsection{Implementation Details}

\par Our experiments utilized the PyTorch framework and an NVIDIA A40 GPU for model training. All networks were trained for 200 epochs using a batch size of 8. The Adam optimizer was employed for optimizing the model, configured with a learning rate of 0.001. Weight decay is applied in the discriminator's optimizer with a value of $1 \times 10^{-5}$. The decay rates for the first and second moments of gradients (denoted as $\beta_1$ and $\beta_2$) were set to 0.9 and 0.999, respectively. The generator's learning rate is dynamically adjusted using the 'ReduceLROnPlateau' scheduler, which monitors the specified metric for improvement. If the metric plateaus for 10 epochs, the scheduler reduces the learning rate by a factor of 0.5, ensuring more efficient optimization in later training stages. We trained and tested the on zhengzhou dataset.

\par The proposed method is comprehensively compared with a variety of deep learning network models, including MTS2ONet \citep{dong2024integrating}, Pix2pix \citep{christovam2021pix2pix}, BicycleGAN \citep{zhu2017toward}, MUNIT \citep{huang2018multimodal}, CycleGAN \citep{zhu2017unpaired}, ResViT \citep{dalmaz2022resvit}, NICE-GAN \citep{chen2020reusing}, and CUT \citep{park2020contrastive}. These models encompass diverse architectures and generative frameworks, representing commonly used benchmarks in the field. 
\par The comparison evaluates performance across multiple metrics, such as Peak signal-to-noise ratio (PSNR), structural similarity index measure (SSIM) \citep{SSIM},  mean absolute error (MAE), root mean square error (RMSE), and Fr\'{e}chet Inception Distance (FID)  \citep{heusel2017gans}. 

\subsection{Quantitative Evaluation}
The experimental results, as summarized in Table \ref{table:result}, highlight the superiority of CRSynthNet over several comparison models. The proposed method achieved the highest PSNR of 26.978 and SSIM of 0.648, demonstrating its capability to generate high-quality outputs with improved structural similarity and signal fidelity. Furthermore, it outperformed comparative models in terms of evaluation metrics, attaining the lowest MAE (0.041) and RMSE (0.050). In addition, the FID score of 72.789 further corroborates the perceptual quality of the generated images.
These results underscore the effectiveness of the proposed method in preserving structural details and ensuring perceptual realism. The notable improvement across all evaluation metrics highlights its potential for optical image synthesis.

\begin{table}[h!]
\centering
\begin{tabular}{|l|c|c|c|c|c|}
\hline
\textbf{Model} & \textbf{PSNR (↑)} & \textbf{SSIM (↑)} & \textbf{MAE (↓)} & \textbf{RMSE (↓)} & \textbf{FID (↓)} \\ \hline
MTS2ONet & 26.225 & 0.622 & 0.049 & 0.057 & 81.150 \\ \hline
CRSynthNet & \textbf{26.978} & \textbf{0.648} & \textbf{0.041} & \textbf{0.050} & \textbf{72.789} \\ \hline
Pix2pix & 20.317 & 0.389 & 0.073 & 0.100 & 142.605 \\ \hline
BicycleGAN & 22.560 & 0.473 & 0.050 & 0.079 & 128.137 \\ \hline
MUNIT & 18.514 & 0.327 & 0.094 & 0.128 & 121.453 \\ \hline
CycleGAN & 18.388 & 0.315 & 0.103 & 0.131 & 114.503 \\ \hline
ResViT & 21.331 & 0.476 & 0.067 & 0.090 & 233.827 \\ \hline
NICE-GAN & 20.206 & 0.395 & 0.079 & 0.108 & 203.176 \\ \hline
CUT & 17.241 & 0.363 & 0.089 & 0.139 & 194.489 \\ \hline
\end{tabular}
\caption{Comparison of different models based on various metrics.}
\label{table:result}
\end{table}

\subsection{Qualitative Evaluation}
\par
\begin{figure}[h!]
    \centering
    \includegraphics[width=\linewidth]{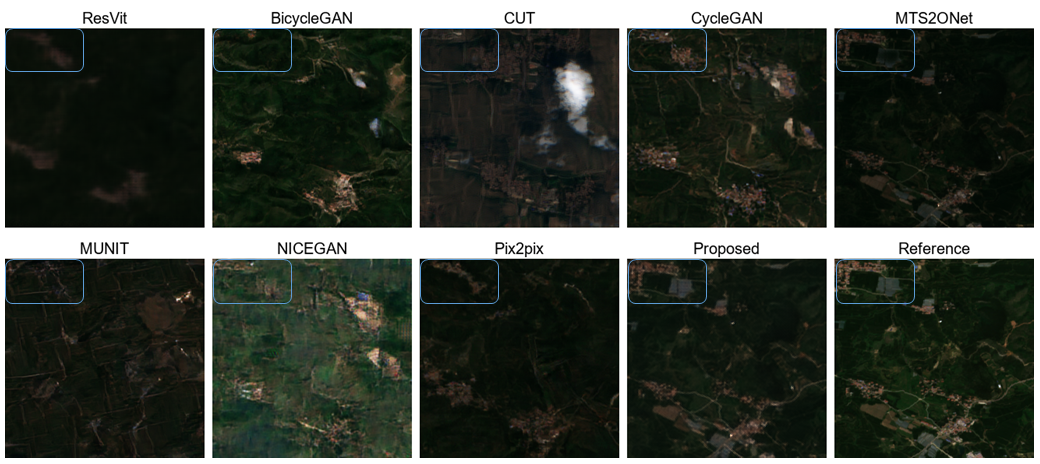} 
    \caption{Generated results for woodland scene from different methods.}
    \label{fig:result_woodland}
\end{figure}

\begin{figure}[h!]
    \centering
    \includegraphics[width=\linewidth]{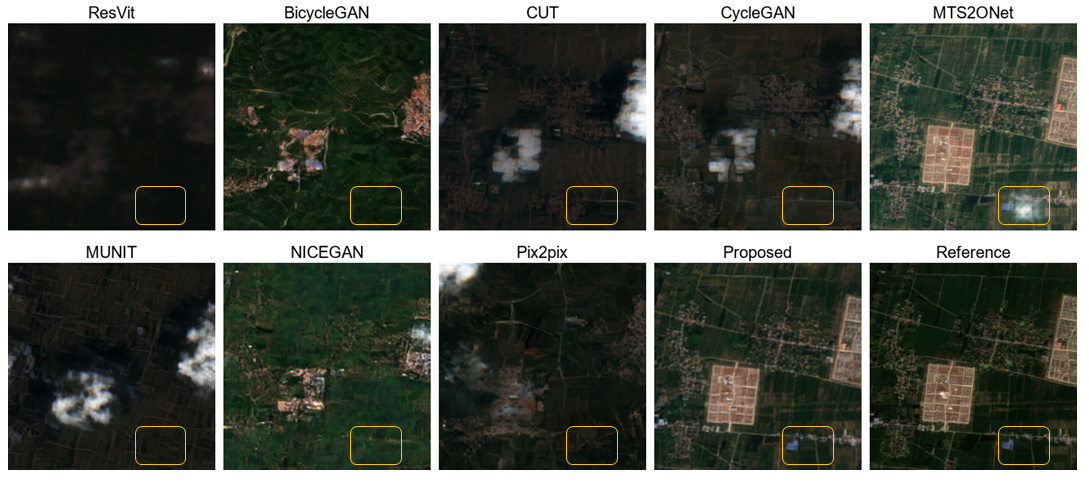} 
    \caption{Generated results for farmland scene from different methods.}
    \label{fig:result_farmland}
\end{figure}

\begin{figure}[h!]
    \centering
    \includegraphics[width=\linewidth]{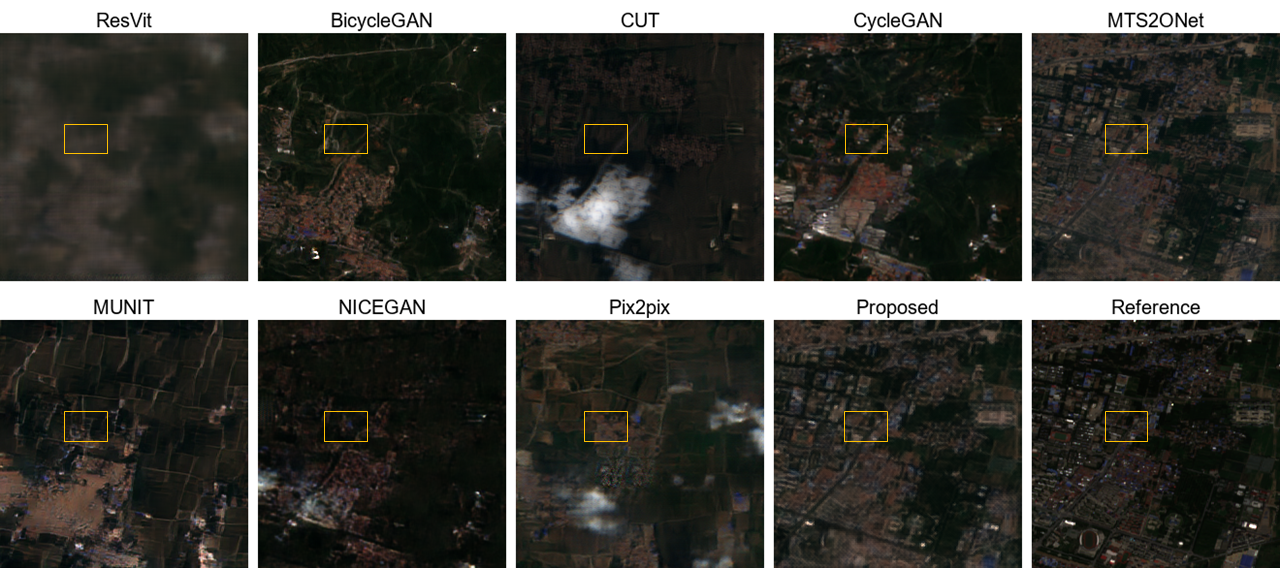} 
    \caption{Generated results for urban building scene from different methods.}
    \label{fig:result_building1}
\end{figure}

\begin{figure}[h!]
    \centering
    \includegraphics[width=\linewidth]{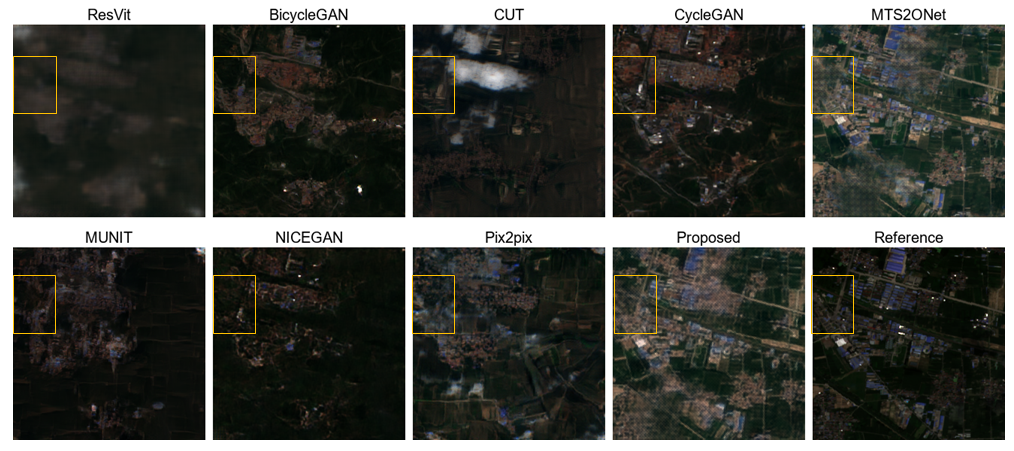} 
    \caption{Generated results for village scene from different methods.}
    \label{fig:result_building2}
\end{figure}

\begin{figure}[h!]
    \centering
    \includegraphics[width=\linewidth]{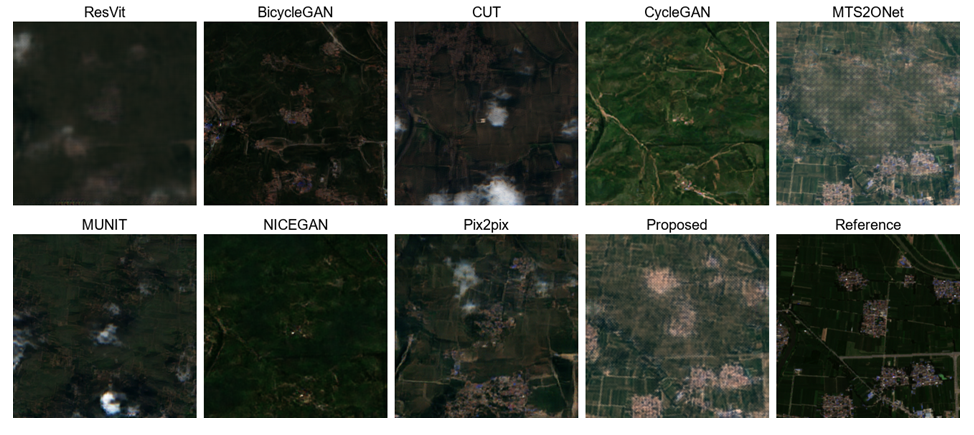} 
    \caption{Generated results for challenging farmland scene from different methods.}
    \label{fig:result_farmland2}
\end{figure}

 In this section, we qualitatively analyze the experimental results by presenting representative images, including woodland scenes, urban building scenes, village, and farmland scenes. The selected scenarios offer valuable examples of the model's applicability and reliability across various environments. Figure \ref{fig:result_woodland} to Figure \ref{fig:result_farmland2} show the generated results from different methods and the reference image. To ensure a clear comparison, the results of MTS2ONet and the proposed method are placed next to the reference image. 

\par Figure \ref{fig:result_woodland} illustrates a woodland scene. It is evident that CRSynthNet's result demonstrates notable detail preservation and visual similarity to the reference image. In the area highlighted by the blue box, there are two distinct regions of vegetation characterized by well-defined and structured boundaries, indicative of human-managed planting. Compared to other results, CRSynthNet demonstrates sharper boundaries in these regions. Additionally, the residential area in the lower part of the image is reconstructed with exceptional detail. Conversely, in the outputs of other methods, the roads within the image are not as clear as the results of CRSynthNet. The road boundaries are poorly defined, indicating less precise reconstruction in those areas.

\par Figure \ref{fig:result_farmland} presents a farmland scene. The results produced by CRSynthNet demonstrate no interference from clouds. Although BicycleGAN's output is similarly unaffected, it fails to reconstruct the farmland scene accurately and instead generates an unrelated image. NICEGAN partially restores the spectral features but lacks detailed reconstruction. For instance, the blue container within the yellow box is not recovered in NICEGAN's results. In contrast, the small and clearly defined house is distinctly and effectively reconstructed by CRSynthNet, highlighting its strong ability to accurately capture fine details.

\par Figure \ref{fig:result_building1} depicts an urban scene characterized by densely packed buildings and complex street networks. It can be observed that MUNIT and BicycleGAN introduce unrealistic features, such as artificial forested and farmland areas, which do not align with the characteristics of the actual urban scene in the reference image. Similarly, CycleGAN reconstructs an area of non-existent red buildings that do not correspond to the actual scene. Outputs from NICEGAN, Pix2Pix, and CUT exhibit significant cloud interference, which visually obscures parts of the reconstructed scenes and affects their overall clarity.
\par At the top of the image, the horizontal road in the reference image is rendered differently by various methods. In BicycleGAN's results, it appears as a woodland path, while in CUT and CycleGAN's outputs, it resembles a waterway cutting through grasslands. Only MTS2ONet and the proposed method successfully reconstruct this artificial build-up road as it should appear. 
\par In the central region highlighted by the yellow box, CRSynthNet demonstrates outstanding performance. It provides clearer road details, more defined building edges, and shapes closely matching the reference image, outperforming other methods in capturing the intricate features of this urban area.

\par Figure \ref{fig:result_building2} depicts a complex village scene, characterized by a broader range of building densities, a greater variety of building colors, and diverse spectral features. This scene also includes farmland areas, which present notable reconstruction challenges due to rich variety of features. In the resulting images, ResVit struggles to produce visually meaningful outputs, falling short in reconstructing the intended scenes. BicycleGAN successfully eliminates cloud interference but does not align with the desired reference image. CUT faces difficulties with heavy cloud occlusions, and even in cloud-free areas, the reconstruction quality remains suboptimal. Pix2Pix performs relatively well by recovering certain spectral features, such as blue-colored structures, but still lacks the fidelity needed to match the reference image. MTS2ONet shows a strong ability to reconstruct structural elements, such as man-made features. An analysis of the yellow box area reveals that CRSynthNet outperforms other methods across multiple dimensions. It excels in reconstructing intricate details such as building contours, and roads. Furthermore, CRSynthNet stands out by effectively tackling complex scenarios, reducing cloud interference, preserving spectral fidelity.

\par Figure \ref{fig:result_farmland2} shows results from a challenging farmland scene image. Among the tested methods, only MTS2ONet and CRSynthNet were able to roughly restore the distinct spectral features of different regions within the image. However, these methods still face challenges in restoring consistent and structured details within the image. The proposed method demonstrates a advantage in reducing the severity of blurry patterns compared to MTS2ONet. Additionally, it effectively addresses the challenge of obscuring clouds in the image. While some gaps remain in replicating the intricate details and spectral precision of the reference image, the proposed method demonstrates better robustness compared to other methods, better at handling complex scenarios under challenging conditions.

\subsection{Ablation Experiments}
\par Ablation experiments were conducted to assess the impact of key components on model performance. Specifically, this section primarily evaluates the effectiveness of the DownUpBlock, FusionAttention, designed discriminator as well as the spatial and channel Attention mechanisms integrated within the decoder. 
\par The absence of the DownUpBlock led to a noticeable performance drop, with a PSNR of 27.053 and an FID of 73.570, reflecting its critical role in enhancing image reconstruction and overall consistency. Similarly, excluding the FusionAttention resulted in a further decline, with a PSNR of 26.396 and a higher FID of 74.758, demonstrating its contribution to fuse the features from two dates. It imroves the network's ability to accurately capture subtle changes and preserve essential details across time variations.
\par The Channel Attention and Spatial Attention modules, when removed, also showed a significant reduction in performance, as these modules are essential for capturing critical features and improving feature selection. The absence of Channel Attention yielded a PSNR of 26.237 and an FID of 71.875, while the removal of Spatial Attention exhibited the weakest performance among all settings, with a PSNR of 25.808 and an FID of 84.289. Channel Attention enhances the representation of spectral features by emphasizing critical channels, improving spectral fidelity. Spatial Attention, however, plays a larger role by directly refining structural details, such as edges and textures, making it more impactful for maintaining spatial integrity in image reconstruction.
\par Compared to the results of using DIS in [13], CRSynthNet, achieves superior performance across most metrics, including an SSIM of 0.648 and a relatively lower FID of 72.789. This demonstrates the designed discriminator providing more precise feedback to the generator during training.
\par Overall, the ablation study confirms that each designed component contributes significantly to the effectiveness of CRSynthNet, with the complete model achieving the most balanced and optimal results across all evaluated metrics.

\begin{table}[h!]
\centering
\caption{Ablation study of the CRSynthNet using five evaluation metrics. Abbreviations: Att = Attention, DIS = Discriminator, DownUp = DownUpBlock. "DIS in \citep{dong2024integrating}" in the table refers to the discriminator in the experiment is proposed in \citep{dong2024integrating}.}
\label{tab:ablation_results}
\begin{tabular}{|l|c|c|c|c|c|}
\hline
\textbf{Setting} & \textbf{PSNR (↑)} & \textbf{SSIM (↑)} & \textbf{MAE (↓)} & \textbf{RMSE (↓)} & \textbf{FID (↓)} \\ \hline
{No\_DownUp} & 27.053 & 0.600 & 0.034 & 0.049 & 73.570 \\ \hline
{No\_FusionAtt}    & 26.396 & 0.632 & 0.046 & 0.056 & 74.758 \\ \hline
{DIS in \citep{dong2024integrating}}    & 26.339 & 0.632 & 0.047 & 0.056 & 78.300 \\ \hline
{No\_Channel Att}     & 26.237 & 0.627 & 0.048 & 0.057 & 71.875 \\ \hline
{No\_Spatial Att}    & 25.808 & 0.614 & 0.051 & 0.060 & 84.289 \\ \hline
{CRSynthNet} & \textbf{26.978} & \textbf{0.648} & \textbf{0.041} & \textbf{0.050} & \textbf{72.789} \\ \hline
\end{tabular}
\end{table}

\section{Discussion}
\par This paper proposed an advanced whole image reconstruction method, CRSynthNet, incorporating novel modules and refined discriminator, to address challenges in satellite image synthesis under complex scenarios.
\subsection{Efficiency of CRSynthNet}
\par The experimental results of CRSynthNet highlight its effectiveness in addressing challenging satellite image synthesis scenarios, particularly better at dealing with cloud-covered temporal images. The integration of novel components, such as the DownUp Block, Fusion Attention, and Spatial and Channel Attention mechanisms, plays a pivotal role in achieving high accuracy.  
\par The DownUp Block is crucial for extracting essential features while preserving spatial resolution, which plays a key role in restoring overall spectral characteristics. Fusion Attention integrates temporal features across different timestamps, ensuring spectral fidelity and addressing temporal inconsistencies. Moreover, it prioritizes crucial regions within the image, emphasizing areas that require detailed reconstruction. This targeted focus enhances the representation of key features, enabling precise restoration of complex structures and regions with significant variations. In the decoder, spatial attention enhances spatial details such as shapes and textures, while channel attention emphasizes inter-channel relationships to improve spectral representation. Additionally, The discriminator enhances adversarial learning by effectively guiding the generator, allowing it to address blurry patterns and accurately reconstruct intricate details. This comprehensive design ensures high accuracy and reliable restoration across complex environments.

\par The designed method achieves remarkable precision, with significant improvements over existing methods. For example, the SSIM value increased from 0.622 to 0.648, while the FID value decreased from 81.150 to 72.789. Most notably, the method demonstrates superior visual performance by effectively mitigating the impact of cloud cover from temporal images and achieving precise restoration of structural details and textures. It excels in reconstructing complex features across diverse urban and rural landscapes, ensuring high fidelity and clarity in the synthesized imagery.

\subsection{Limitations} 
\par While the proposed method demonstrates strong performance in satellite image synthesis, it faces challenges in certain scenario, particularly when the missing features are underrepresented in optical images. As shown in Figure \ref{fig:result_farmland2}, small building areas are covered by clouds and the whole image is dominated by farmland. In such cases, the cloud-free regions of optical images fail to provide sufficient structural and detail information necessary for reconstructing the building area. Furthermore, SAR images record electromagnetic scattering properties rather than the spectral and spatial details. This restricts their ability to supplement the fine textures and intricate patterns. 

\section{Conclusion}
\par This paper proposes a novel image synthesis network, CRSynthNet, designed to address challenges in reconstructing missing optical images in complex scenarios. By incorporating the DownUp Block, Fusion Attention, and Decoder, CRSynthNet significantly improves satellite image reconstruction. It effectively restores structural details, such as edges and textures, while preserving spectral features. Additionally, the refined discriminator enhances adversarial learning, enabling the generation of more realistic and coherent results. Experimental results confirm the effectiveness of this method, with ablation studies highlighting the importance of each component in enhancing performance.
\par In addition to the methodological advancements, this paper contributes a newly developed dataset, designed to address specific cloud cover challenges in satellite image synthesis. This dataset serves as a valuable resource for the research community, facilitating further exploration in this domain. Unlike earlier datasets that rely on cloud-free complementary optical images, the TCSEN12 dataset includes optical images with cloud cover from Time 1 to better align with real-world scenarios. Furthermore, unlike datasets that use later timestamps to reconstruct earlier images, the TCSEN12 dataset utilizes earlier timestamps to predict and reconstruct images from subsequent times. The reference images contain less than 5\% cloud coverage, ensuring that the target images used for training are nearly cloud-free and align with real-world scenarios. The dataset includes images from the Zhengzhou flood region, characterized by temporal land cover change. This diversity provides a robust testbed for evaluating the generalization capability of various methods, making the dataset highly suitable for advancing satellite image systhnsis research.
\par Overall, this work makes significant contributions to the field by proposing a robust solution for handling challenging scenarios and introducing a dataset that fosters further advancements. Future research could focus on improving multi-modal feature fusion and integrating additional auxiliary data to enhance synthesis quality in feature-sparse environments with limited representation in optical images.

\section*{Code availability}
The data partition documents and the code implemented in this study will be available at: https://github.com/chenxiduan/MultiTemporalCloudFree.

\section*{data availability statement}
The data that support the findings of this study are openly available at https://github.com/chenxiduan/MultiTemporalCloudFree.

\section*{Disclosure statement}
\par The authors declare that they have no known competing financial interests or personal relationships that could have appeared to influence the work reported in this paper.

\bibliographystyle{elsarticle-num}
\bibliography{abb}

\begin{thebibliography}{10}
\expandafter\ifx\csname url\endcsname\relax
  \def\url#1{\texttt{#1}}\fi
\expandafter\ifx\csname urlprefix\endcsname\relax\def\urlprefix{URL }\fi
\expandafter\ifx\csname href\endcsname\relax
  \def\href#1#2{#2} \def\path#1{#1}\fi

\bibitem{nguyen2022improvement}
T.~H. Nguyen, S.~Ricci, C.~Fatras, A.~Piacentini, A.~Delmotte, E.~Lavergne, P.~Kettig, Improvement of flood extent representation with remote sensing data and data assimilation, IEEE Transactions on Geoscience and Remote Sensing 60 (2022) 1--22.

\bibitem{yang2021evolution}
L.~Yang, Q.~Guan, J.~Lin, J.~Tian, Z.~Tan, H.~Li, Evolution of ndvi secular trends and responses to climate change: A perspective from nonlinearity and nonstationarity characteristics, Remote sensing of environment 254 (2021) 112247.

\bibitem{wang2022unetformer}
L.~Wang, R.~Li, C.~Zhang, S.~Fang, C.~Duan, X.~Meng, P.~M. Atkinson, Unetformer: A unet-like transformer for efficient semantic segmentation of remote sensing urban scene imagery, ISPRS Journal of Photogrammetry and Remote Sensing 190 (2022) 196--214.

\bibitem{SONG2024104932}
Y.~Song, D.~Gnyawali, L.~Qian, From early curiosity to space wide web: The emergence of the small satellite innovation ecosystem, Research Policy 53~(2) (2024) 104932.
\newblock \href {http://dx.doi.org/https://doi.org/10.1016/j.respol.2023.104932} {\path{doi:https://doi.org/10.1016/j.respol.2023.104932}}.

\bibitem{gonzalezcalabuig2025generative}
M.~Gonzalez-Calabuig, M.~Ángel Fernández-Torres, G.~Camps-Valls, Generative networks for spatio-temporal gap filling of sentinel-2 reflectances, ISPRS Journal of Photogrammetry and Remote Sensing 220 (2025) 637--648.
\newblock \href {http://dx.doi.org/https://doi.org/10.1016/j.isprsjprs.2025.01.016} {\path{doi:https://doi.org/10.1016/j.isprsjprs.2025.01.016}}.

\bibitem{zhu2022urban}
W.~Zhu, Z.~Cao, P.~Luo, Z.~Tang, Y.~Zhang, M.~Hu, B.~He, Urban flood-related remote sensing: research trends, gaps and opportunities, Remote Sensing 14~(21) (2022) 5505.

\bibitem{christovam2021pix2pix}
L.~E. Christovam, M.~H. Shimabukuro, M.~d. L.~B. Galo, E.~Honkavaara, Pix2pix conditional generative adversarial network with mlp loss function for cloud removal in a cropland time series, Remote Sensing 14~(1) (2021) 144.

\bibitem{li2024improved}
Y.~Li, Q.~Liu, S.~Chen, X.~Zhang, An improved gap-filling method for reconstructing dense time-series images from landsat 7 slc-off data, Remote Sensing 16~(12) (2024) 2064.

\bibitem{zeng2020towards}
Z.~Zeng, Y.~Gan, A.~J. Kettner, Q.~Yang, C.~Zeng, G.~R. Brakenridge, Y.~Hong, Towards high resolution flood monitoring: An integrated methodology using passive microwave brightness temperatures and sentinel synthetic aperture radar imagery, Journal of Hydrology 582 (2020) 124377.

\bibitem{sun2017haze}
L.~Sun, R.~Latifovic, D.~Pouliot, Haze removal based on a fully automated and improved haze optimized transformation for landsat imagery over land, Remote Sensing 9~(10) (2017) 972.

\bibitem{xu2022attention}
M.~Xu, F.~Deng, S.~Jia, X.~Jia, A.~J. Plaza, Attention mechanism-based generative adversarial networks for cloud removal in landsat images, Remote sensing of environment 271 (2022) 112902.

\bibitem{yang2022sar}
X.~Yang, J.~Zhao, Z.~Wei, N.~Wang, X.~Gao, Sar-to-optical image translation based on improved cgan, Pattern Recognition 121 (2022) 108208.

\bibitem{dong2024integrating}
C.~Dong, G.~Yang, Y.~Wang, W.~Sun, X.~Meng, B.~Chen, Integrating multi-temporal sar and optical information for missing optical imagery generation, IEEE Transactions on Geoscience and Remote Sensing.

\bibitem{addink1999comparison}
E.~Addink, A.~Stein, A comparison of conventional and geostatistical methods to replace clouded pixels in noaa-avhrr images, International Journal of Remote Sensing 20~(5) (1999) 961--977.

\bibitem{van2012remote}
F.~Van~der Meer, Remote-sensing image analysis and geostatistics, International Journal of Remote Sensing 33~(18) (2012) 5644--5676.

\bibitem{lin2012cloud}
C.-H. Lin, P.-H. Tsai, K.-H. Lai, J.-Y. Chen, Cloud removal from multitemporal satellite images using information cloning, IEEE transactions on geoscience and remote sensing 51~(1) (2012) 232--241.

\bibitem{zhang2018missing}
Q.~Zhang, Q.~Yuan, C.~Zeng, X.~Li, Y.~Wei, Missing data reconstruction in remote sensing image with a unified spatial--temporal--spectral deep convolutional neural network, IEEE Transactions on Geoscience and Remote Sensing 56~(8) (2018) 4274--4288.

\bibitem{duan2020thick}
C.~Duan, J.~Pan, R.~Li, Thick cloud removal of remote sensing images using temporal smoothness and sparsity regularized tensor optimization, Remote Sensing 12~(20) (2020) 3446.

\bibitem{chen2024thick}
Y.~Chen, M.~Chen, W.~He, J.~Zeng, M.~Huang, Y.-B. Zheng, Thick cloud removal in multitemporal remote sensing images via low-rank regularized self-supervised network, IEEE Transactions on Geoscience and Remote Sensing 62 (2024) 1--13.

\bibitem{zhang2021combined}
Q.~Zhang, Q.~Yuan, Z.~Li, F.~Sun, L.~Zhang, Combined deep prior with low-rank tensor svd for thick cloud removal in multitemporal images, ISPRS Journal of Photogrammetry and Remote Sensing 177 (2021) 161--173.

\bibitem{wang2025mst}
L.~Wang, Q.~Wang, X.~Tong, P.~M. Atkinson, Mst-net: A general deep learning model for thick cloud removal from optical images, IEEE Transactions on Geoscience and Remote Sensing 63 (2025) 1--18.
\newblock \href {http://dx.doi.org/10.1109/TGRS.2025.3543617} {\path{doi:10.1109/TGRS.2025.3543617}}.

\bibitem{zhou2020reconstruction}
F.~Zhou, D.~Zhong, R.~Peiman, Reconstruction of cloud-free sentinel-2 image time-series using an extended spatiotemporal image fusion approach, Remote Sensing 12~(16) (2020) 2595.

\bibitem{merkle2017possibility}
N.~Merkle, P.~Fischer, S.~Auer, R.~M{\"u}ller, On the possibility of conditional adversarial networks for multi-sensor image matching, in: 2017 IEEE International Geoscience and Remote Sensing Symposium (IGARSS), IEEE, 2017, pp. 2633--2636.

\bibitem{fuentes2019sar}
M.~Fuentes~Reyes, S.~Auer, N.~Merkle, C.~Henry, M.~Schmitt, Sar-to-optical image translation based on conditional generative adversarial networks—optimization, opportunities and limits, Remote Sensing 11~(17) (2019) 2067.

\bibitem{wang2019sar}
L.~Wang, X.~Xu, Y.~Yu, R.~Yang, R.~Gui, Z.~Xu, F.~Pu, Sar-to-optical image translation using supervised cycle-consistent adversarial networks, Ieee Access 7 (2019) 129136--129149.

\bibitem{doi2020gan}
K.~Doi, K.~Sakurada, M.~Onishi, A.~Iwasaki, Gan-based sar-to-optical image translation with region information, in: IGARSS 2020-2020 IEEE International Geoscience and Remote Sensing Symposium, IEEE, 2020, pp. 2069--2072.

\bibitem{guo2021edge}
J.~Guo, C.~He, M.~Zhang, Y.~Li, X.~Gao, B.~Song, Edge-preserving convolutional generative adversarial networks for sar-to-optical image translation, Remote Sensing 13~(18) (2021) 3575.

\bibitem{meraner2020cloud}
A.~Meraner, P.~Ebel, X.~X. Zhu, M.~Schmitt, Cloud removal in sentinel-2 imagery using a deep residual neural network and sar-optical data fusion, ISPRS J. Photogramm. Remote. Sens. 166 (2020) 333--346.

\bibitem{xu2022glf}
F.~Xu, Y.~Shi, P.~Ebel, L.~Yu, G.-S. Xia, W.~Yang, X.~X. Zhu, Glf-cr: Sar-enhanced cloud removal with global--local fusion, ISPRS J. Photogramm. Remote. Sens. 192 (2022) 268--278.

\bibitem{mao2022cloud}
R.~Mao, H.~Li, G.~Ren, Z.~Yin, Cloud removal based on sar-optical remote sensing data fusion via a two-flow network, IEEE J. Sel. Top. In Appl. Earth Obs. Remote. Sens. 15 (2022) 7677--7686.

\bibitem{duan2024feature}
C.~Duan, M.~Belgiu, A.~Stein, Feature enhancement network for cloud removal in optical images by fusing with sar images, Int. J. Remote. Sens. 45~(1) (2024) 51--67.

\bibitem{xu2023multimodal}
F.~Xu, Y.~Shi, P.~Ebel, W.~Yang, X.~X. Zhu, Multimodal and multiresolution data fusion for high-resolution cloud removal: A novel baseline and benchmark, IEEE Transactions on Geoscience and Remote Sensing 62 (2023) 1--15.

\bibitem{liu2022swin2}
Z.~Liu, H.~Hu, Y.~Lin, Z.~Yao, Z.~Xie, Y.~Wei, J.~Ning, Y.~Cao, Z.~Zhang, L.~Dong, F.~Wei, B.~Guo, Swin transformer v2: Scaling up capacity and resolution, in: 2022 IEEE/CVF Conference on Computer Vision and Pattern Recognition (CVPR), 2022, pp. 11999--12009.
\newblock \href {http://dx.doi.org/10.1109/CVPR52688.2022.01170} {\path{doi:10.1109/CVPR52688.2022.01170}}.

\bibitem{mao2017least}
X.~Mao, Q.~Li, H.~Xie, R.~Y. Lau, Z.~Wang, S.~Paul~Smolley, Least squares generative adversarial networks, in: Proceedings of the IEEE international conference on computer vision, 2017, pp. 2794--2802.

\bibitem{simonyan2015a}
K.~Simonyan, A.~Zisserman, Very deep convolutional networks for large-scale image recognition, Computational and Biological Learning Society, 2015, pp. 1--14.

\bibitem{MS-SSIM}
Z.~Wang, E.~Simoncelli, A.~Bovik, Multiscale structural similarity for image quality assessment, in: The Thrity-Seventh Asilomar Conference on Signals, Systems \& Computers, 2003, Vol.~2, 2003, pp. 1398--1402 Vol.2.
\newblock \href {http://dx.doi.org/10.1109/ACSSC.2003.1292216} {\path{doi:10.1109/ACSSC.2003.1292216}}.

\bibitem{SSIM}
Z.~Wang, A.~Bovik, H.~Sheikh, E.~Simoncelli, Image quality assessment: from error visibility to structural similarity, IEEE Transactions on Image Processing 13~(4) (2004) 600--612.
\newblock \href {http://dx.doi.org/10.1109/TIP.2003.819861} {\path{doi:10.1109/TIP.2003.819861}}.

\bibitem{pmlr-v70-arjovsky17a}
M.~Arjovsky, S.~Chintala, L.~Bottou, {W}asserstein generative adversarial networks, in: D.~Precup, Y.~W. Teh (Eds.), Proceedings of the 34th International Conference on Machine Learning, Vol.~70 of Proceedings of Machine Learning Research, PMLR, 2017, pp. 214--223.

\bibitem{zhu2017toward}
J.-Y. Zhu, R.~Zhang, D.~Pathak, T.~Darrell, A.~A. Efros, O.~Wang, E.~Shechtman, Toward multimodal image-to-image translation, Advances in neural information processing systems 30.

\bibitem{huang2018multimodal}
X.~Huang, M.-Y. Liu, S.~Belongie, J.~Kautz, Multimodal unsupervised image-to-image translation, in: Proceedings of the European conference on computer vision (ECCV), 2018, pp. 172--189.

\bibitem{zhu2017unpaired}
J.-Y. Zhu, T.~Park, P.~Isola, A.~A. Efros, Unpaired image-to-image translation using cycle-consistent adversarial networks, in: Proceedings of the IEEE international conference on computer vision, 2017, pp. 2223--2232.

\bibitem{dalmaz2022resvit}
O.~Dalmaz, M.~Yurt, T.~{\c{C}}ukur, Resvit: residual vision transformers for multimodal medical image synthesis, IEEE Transactions on Medical Imaging 41~(10) (2022) 2598--2614.

\bibitem{chen2020reusing}
R.~Chen, W.~Huang, B.~Huang, F.~Sun, B.~Fang, Reusing discriminators for encoding: Towards unsupervised image-to-image translation, in: Proceedings of the IEEE/CVF conference on computer vision and pattern recognition, 2020, pp. 8168--8177.

\bibitem{park2020contrastive}
T.~Park, A.~A. Efros, R.~Zhang, J.-Y. Zhu, Contrastive learning for unpaired image-to-image translation, in: Computer Vision--ECCV 2020: 16th European Conference, Glasgow, UK, August 23--28, 2020, Proceedings, Part IX 16, Springer, 2020, pp. 319--345.

\bibitem{heusel2017gans}
M.~Heusel, H.~Ramsauer, T.~Unterthiner, B.~Nessler, S.~Hochreiter, Gans trained by a two time-scale update rule converge to a local nash equilibrium, Advances in neural information processing systems 30.

\end{thebibliography}

\end{document}